\documentclass{article}
\usepackage{spconf,amsmath,graphicx,float,cite}
\usepackage{url}
\pdfoutput=1

\title{Image Tag Completion by Low-rank Factorization with Dual Reconstruction Structure Preserved}

\name{Xue Li $^a$, Yu-Jin Zhang $^a$, Bin Shen $^b$, Bao-Di Liu $^c$}
\address{$^a$ Electronic Engineering, Tsinghua University,	Beijing, 100084, China\\
         $^b$ Computer Science, Purdue University, West Lafayette, IN 47907, USA\\
         $^c$ Information and Control Engineering, China University of Petroleum, Qingdao, 266580, China\\
	xue-li11@mails.tainghua.edu.cn, zhang-yj@mail.tsinghua.edu.cn
        \thanks{This work was supported by National Nature Science Foundation (NNSF: 61171118) and
Specialized Research Fund for the Doctoral Program of Higher Education (SRFDP-20110002110057).}}

\begin{document}
\ninept

\maketitle

\begin{abstract}
A novel tag completion algorithm is proposed in this paper, which is designed with
the following features:
1) {\it Low-rank and error sparsity}:
the incomplete initial tagging matrix $D$ is decomposed into the complete tagging
matrix $A$ and a sparse error matrix $E$. However, instead of minimizing its nuclear
norm, $A$ is further factorized into a basis matrix $U$ and a sparse coefficient
matrix $V$, {\it i.e.} $D=UV+E$. This low-rank formulation encapsulating sparse coding
enables our algorithm to recover latent structures from noisy initial data and avoid
performing too much denoising;
2) {\it Local reconstruction structure consistency}:
to steer the completion of $D$, the local linear reconstruction structures in
feature space and tag space are obtained and preserved by $U$ and $V$ respectively.
Such a scheme could alleviate the negative effect of distances measured by low-level
features and incomplete tags. Thus, we can seek a balance between exploiting
as much information and not being mislead to suboptimal performance.
Experiments conducted on Corel5k dataset and the newly issued Flickr30Concepts dataset
demonstrate the effectiveness and efficiency of the proposed method.
\end{abstract}
\section{Introduction}
\label{sec:intro}
With digital imaging gains its popularity in recent decades, the demand for
effective and efficient automatic image annotation (AIA) methods is highlighted by
both content based image retrieval (CBIR) \cite{zheng2013lp,zheng2014bayes,zheng2014coupled}
and tag based image retrieval (TBIR).
Nevertheless, performance of most existing AIA \cite{makadia2008new,guillaumin2009tagprop,zhengtopic}
methods degrades dramatically when initial tags are
noisy or incomplete, thus how to perform accurate tag completion has become a hot
issue that needs to be addressed.

Among the various proposed methods\cite{zhu2010image,wu2013tag,lin2013image,liu2012image,lee2010map,liu2011image,liu2010image}
for tag completion, the pursuit of
maintaining content consistency and tag relationship has always been a key
component in nearly every algorithm, though in different formulations.
G. Zhu {\it et al.}\cite{zhu2010image} defined two similarity matrix in both
feature space and tag space, and violations of such similarity resulted from the
completed matrix are minimized. Similarly, in \cite{liu2012image},
X.Liu {\it et al.} promoted feature-label harmoniousness and punished interlabel
discrepancy. The recently proposed TMC method \cite{wu2013tag} aimed at preserving
correlation structures for images and tags in the completed matrix, and the LSR
method \cite{lin2013image} performed linear sparse reconstruction for each image and
each tag, respectively. According to their reported performance, LSR is better than
existing methods, especially the ones defining similarity based on distance in feature
space or initial tags, since similar features do not necessarily guarantee related
tags due to the semantic gap, and distances measured by incomplete tags are
unreliable. Therefore, the usage of such distances may introduce risks and mislead
the completion process. On the other hand, analogous to Local Linear Embedding (LLE)
\cite{zheng2011graph}, the LSR method restricts the related images or tags to be
within the same subspace and preserves local geometry, which means the noisy distances
are not involved in this framework, thus the influences of semantic gap and incomplete tags
get alleviated.

Another debated issue involves the low-rank constraint. As pointed out in
\cite{zhu2010image}, low-rank constraint is natural since the semantic space spanned
by tags is low-rank, whereas \cite{lin2013image} indicates that the low-rank constraint
may be more suitable for denoising rather than completion, and it is difficult to
control the degree of denoising. However, methods that do not utilize low-rank property
strongly rely on initial tags, since they lack the ability to recover latent
structures with noisy incomplete data.

Motivated by the foregoing analysis, our formulation is designed with the following features:
\begin{itemize}
\item {\bf Low-rank and error sparsity.} The initial tagging matrix $D$ is decomposed
into a sparse error matrix $E$ and a factorization of a basis matrix $U$ and a sparse
coefficient matrix $V$, {\it i.e.} $D=UV+E$.
This low-rank  formulation encapsulating sparse coding \cite{liu2012discriminant,liu2012learning,shen2009image,shen2010non,wu2013visual} has the ability to
recover the latent complete matrix from noisy data and at the same time avoid performing
too much denoising, which is a main problem of \cite{zhu2010image}.
\item {\bf Local reconstruction structure consistency.} As discussed above, using distances
measured by low-level features and incomplete tags may introduce risks and mislead the
completion process. Therefore, similar to \cite{lin2013image}, the proposed method also
rests on the LLE assumption and attempts to preserve the local linear reconstruction
structures in {\sl both} the feature space and tag space.
\end{itemize}

The main contribution of the proposed formulation lies in the combination and extension of
low-rank property and local reconstruction structure consistency.
For the former, despite its ability to recover low-rank structures from noisy data, minimizing
nuclear norm is more suitable for the reconstruction of dense matrix, thus it may tend to perform
filtering rather than completion, and the recovered tags may not be accurate enough, especially
when $A$ itself is sparse. In order to fix this, the proposed method uses sparse coding for the
reconstruction of the low-rank final matrix $A$, which can then seek a balance between accurate
reconstruction and robustness towards initial noisy data;
For the latter, as an extension of \cite{lin2013image}, the local geometry structures are preserved
in the compressed low-dimensional feature space and tag space, which is more suitable for our low-rank
framework and to steer the generation of $U$ and $V$. Note that the local geometry structures in
the original space is guaranteed to be preserved in this way.

The rest of this paper is organized as follows. The novel formulation for tag completion
is elaborated in Section \ref{sec:tag}, followed by detailed optimization methods
in Section \ref{sec:opt}. Experimental results on two datasets are presented in Section
\ref{sec:exp}, and Section \ref{sec:conclusion} concludes this paper.

\section{Tag Completion by low-rank factorization with dual local reconstruction structure preserved}
\label{sec:tag}

Denote the initial user-provided tagging matrix as $D_{N \times M}$, with $M$
and $N$ specifying the number of tags and images, respectively. Entries in $D$
have binary values, that is,
\begin{align}
    D_{ij} = \left\{
               \begin{array}{ll}
                 1, & \textrm{ in case image } i \textrm{ is associated with
label } j;\\
                 0, & \textrm{ otherwise}.
               \end{array}
             \right.
\end{align}

Our goal is to recover the latent complete tagging matrix $A$. Following the
framework in \cite{zhu2010image}, $D$ is also decomposed into the complete matrix
$A$ and a sparse error matrix $E$; Since $A$ is believed to be of low-rank, then it can
be further factorized as $A = U V$.
Thus,
\begin{eqnarray}\label{eqn:decompose}
  D = U V + E
\end{eqnarray}
where $U_{N \times K}$ and $V_{K \times M}$ are the basis and sparse coefficient
matrix, respectively.

As mentioned in Section \ref{sec:intro}, low-rank is achieved using a sparse coding scheme,
and preserve local reconstruction structures in the compressed low-dimensional feature
space and tag space. Details of the proposed method is presented in the following subsections.

\subsection{Low Rank and Error Sparsity}
\label{ssec:lowrank}

In order to make our method robust to initial noisy labels, the proposed method follows the framework
developed in \cite{zhu2010image} and adopts the low-rank constraint. However, in order
to fix the problem pointed out by \cite{lin2013image}, the complete
tagging matrix $A$ is factorized as $A = U V$, instead of minimizing its nuclear norm. Here $V$ can
be viewed as tag representation in new low-dimensional tag space, and $U$ as image
representation in new low-dimensional image space.
Thus, our basic objective function can be written as follows:
\begin{eqnarray}\label{eqn:obj.function1}
 &&\min_{U,V,E} \:\:
    \left\{
        \big\| D - E - UV \big\|^{2}_{F} + 2\eta \big\|V\big\|_{1} + \beta \big\|E \big\|_{1}
    \right\}\nonumber\\
    \textrm{s.t.} && \|U_{\bullet k}\|_{2}=1, \: \forall k \in {1,2,\cdots,K}
\end{eqnarray}

Note that Eq.(\ref{eqn:obj.function1}) can be interpreted as sparse coding,
with $U_{N \times K}$ being the basis matrix and $V_{K\times M}$ the sparse coefficient
matrix.
Such a scheme can achieve fine-grained approximation and control the degree of denoising,
which makes it more suitable for completion tasks. The error matrix $E$ also has the
ability to prevent it from completely reconstructing $D$.
\subsection{Local Reconstruction Structure in Feature Space}
\label{ssec:feature}

Denote $X_{N \times L}$ as the feature matrix in the original space, each row of
$X$ is a feature vector of an image. In the new low-dimensional space,
each row of $U$ is a compressed representation of an image. Similar to the idea of
LLE, the local geometry structure is believed to be important and should be preserved
while compressing the representation. Thus, first the original data $X$ is explored for
the structure information, which is encoded in matrix $S$:
\begin{eqnarray}\label{eqn:s}
  && S^* = \arg\min_{S}
  \left\{
        \big\|X - SX\big\|^{2}_{F} + \alpha \big\|S\big\|_{1}
  \right\} \nonumber\\
 \textrm{s.t.} && S_{nn}=0, \: \forall n \in {1,2,\cdots,N}
\end{eqnarray}
where  $S_{N \times N}$ is the local linear reconstruction coefficient matrix in feature space.

The $j$-th row of $S$ contains corresponding weights that can be used to reconstruct
the features of the $j$-th image using that of other images.

Eq.(\ref{eqn:s}) can be efficiently solved using the feature-sign method \cite{lee2006efficient}.

Next, assume the tags of the $j$-th image can be equally reconstructed by the tags of other
images, thus $A\sim SA$. The local linear reconstruction structure specified by $S$
should be robust to the sparse coding procedure in Eq.(\ref{eqn:obj.function1}), which
means this reconstruction structure should applies to $U$ as well, {\it i.e.} $U \sim SU$.
Therefore, the objective function can be rewritten as:
\begin{eqnarray}\label{eqn:obj.function2}
&&\min_{U,V,E} \:\:
    \Big\{
        \big\|D-E-UV\big\|^{2}_{F} + \gamma \big\|U-SU\big\|^{2}_{F} \nonumber\\
        && \hspace{1cm} + 2\eta \big\|V\big\|_{1}+\beta \big\|E\big\|_{1}
    \Big\}\nonumber\\
    \textrm{s.t.} && \|U_{\bullet k}\|_{2}=1, \: \forall k \in {1,2,\cdots,K}
\end{eqnarray}

\subsection{Local Reconstruction Structure in Tag Space}
\label{ssec:subhead}
Similarly, each column of $V$ can be viewed as compressed feature of a tag, and the local
reconstruction structure in the original tag space should be preserved. So, the structure
information, encoded in $T$, is explored first in the original data $D$:
\begin{eqnarray}\label{eqn:t}
&&T^* = \arg\min_{T}
      \left\{
            \big\|D-DT\big\|^{2}_{F}+\mu \big\|T\big\|_{1}
      \right\} \nonumber\\
\textrm{s.t.} && T_{mm}=0, \: \forall m \in {1,2,\cdots,M}
\end{eqnarray}
where  $T_{M \times M}$ is the local linear reconstruction coefficient matrix in tag space.

The $i$-th column of $T$ contains corresponding weights that can be used to reconstruct
the distribution of the $i$-th tag using that of other tags.

Then the reconstruction relationship specified by $T$ should also applies to $V$.
Therefore, our final objective function is as follows:
\begin{eqnarray}\label{eqn:final.obej.func}
&&\min_{U,V,E} \:\:
\Big\{
    \big\|D-E-UV\big\|^{2}_{F} + \gamma \big\|U-SU\big\|^{2}_{F} + \nonumber\\
    && \hspace{1cm} \lambda \big\|V-VT\big\|^{2}_{F}+2\eta \big\|V\big\|_{1} + \beta \big\|E\big\|_{1}
\Big\}\nonumber\\
\textrm{s.t.} && \|U_{\bullet k}\|_{2}=1, \: \forall k \in {1,2,\cdots,K}
\end{eqnarray}

Eq.(\ref{eqn:t}) can be solved analogous to Eq.(\ref{eqn:s}).

\section{Optimization}
\label{sec:opt}

In this section, we focus on solving the minimization of the proposed objective function
in Eq.(\ref{eqn:final.obej.func}). Although it is not jointly convex in all three variables,
it is separately convex in $U$, $V$ and $E$ with remaining variables fixed. Thus,
Eq.(\ref{eqn:final.obej.func}) can be solved by decoupling it into three subproblems and
conducting optimization separately.

\subsection{Optimizing Coefficient V}
\label{ssec:opt.v}

Here the method in \cite{liu2012learning} is used. Define $\widetilde{D}=D-E$,
$H=\lambda (T-I)(T-I)^{T}$, when $U$ and $E$ are kept fixed, Eq.(\ref{eqn:final.obej.func}) reduces to:
\begin{eqnarray}\label{eqn:opt.v1}
f(V)&=&\big\|\widetilde{D}-UV\big\|^{2}_{F}+\lambda \big\|V-VT\big\|^{2}_{F}+2\eta \big\|V\big\|_{1}\nonumber\\
&=&tr\big\{\widetilde{D}^{T}\widetilde{D}-2V\widetilde{D}^{T}U+VV^{T}U^{T}U\big\}\nonumber\\
&&+tr\big\{VHV^{T}\big\}+2\eta \big\|V\big\|_{1}\
\end{eqnarray}

Ignoring the constant term $tr\big\{\widetilde{D}^{T}\widetilde{D}\big\}$, the objective
function of $V_{km}$ reduces to
\begin{align}\label{eqn:opt.vkn}
  f(V_{km})
 = & \:2V_{km}\left[ \sum_{\substack{l=1\\l\not =k}}^{K} V_{lm}(U^{T}U)_{lk}+\sum_{\substack{r=1\\r\not=m}}^{M} H_{mr} V_{kr} -(\widetilde{D}^{T}U)_{mk} \right] \nonumber\\
& \hspace{0.5cm} + V_{km}^{2}\big[(U^{T}U)_{kk}+ H_{mm}\big] + 2\eta |V_{km}|.
\end{align}

Note that Eq.(\ref{eqn:opt.vkn}) is a piece-wise parabolic function that opens up, which is
convex and easy to obtain the optimal point
\begin{eqnarray}
V_{km}=\frac{max\{P_{km},\eta\}+min\{P_{km},-\eta\}}{\big(U^{T}U)_{kk}+H_{mm}}
\end{eqnarray}
where
$$P_{km} = (\widetilde{D}^{T}U)_{mk} - \sum_{\substack{l=1\\l\not =k}}^{K} V_{lm}(U^{T}U)_{lk}
- \sum_{\substack{r=1\\r\not=m}}^{M} H_{mr} V_{kr}$$

\subsection{Optimizing Basis U}
\label{ssec:opt.u}
Optimization of $U$ can be conducted by alternating between a procedure similar to $V$ and
Euclidean projection.

Define $G=\gamma (S-I)^{T}(S-I)$, when $V$ and $E$ are fixed, $U$ can be solved analogous to $V$,
the only modification is to remove the $L_{1}$ regularizer:
\begin{eqnarray}
U_{nk} = \frac{Q_{nk}}{(VV^{T})_{kk} + G_{nn}}
\end{eqnarray}
where$$Q_{nk} = (V\widetilde{D}^{T})_{kn} - \sum_{\substack{l=1\\l\not=k}}^{K} (VV^{T})_{kl}U_{nl} - \sum_{\substack{r=1\\r\not=n}}^{N} U_{rk}G_{rn}$$

Then, Euclidean projection is performed to ensure the $L_{2}$ norm of each column in $U$ is
less than $1$. Note this is coordinate descend approach and the projection is conducted after each coordinate is
updated if the $L_{2}$ norm of the updated column of $U$ is greater than 1. Thus, both convergence
and the decrease in objective function are guaranteed. This constraint of $\|U_{\bullet k}\|=1$ is
relaxed to $\|U_{\bullet k}\|\leq 1$, since the relaxation will result in a convex optimization
problem while keeping the global optimum unchanged. {\it i.e.} the optimal $U$ will always satisfy
$\|U_{\bullet k}\|=1$ even if our explicit constraint is $\|U_{\bullet k}\|\leq 1$.
\subsection{Optimizing Sparse Error E}
\label{ssec:subhead}

Finally, when $U$ and $V$ are fixed, obtaining $E$ reduces to solving the following sparse coding
problem:
\begin{eqnarray}
E^* = \arg\min_{E} \:\: \left\{ \big\|D-UV-E\big\|^{2}_{F} + \beta \big\|E\big\|_{1} \right\},
\end{eqnarray}
which can be solved similar to $S$ and $T$.

\subsection{Implementation Issues}
\label{ssec:subhead}

A kNN (k Neareat Neighbors) strategy is adopted when calculating matrix $S$ and $T$, where
$k=200$ (same to \cite{lin2013image}), in order to make it faster. For the number
of basis, $K=100$ is used in Corel5k dataset, and $K=500$ for the much larger Flickr30Concepts dataset.

Meanwhile, similar to \cite{zhu2010image}, $D$ is re-initialized as $D=(SD+DT)/2$ before
fed to the completion process.

Also, for the Flickr30Concepts dataset, tags are treated as features when obtaining $S$.
Tags are not used as features in Corel5k dataset, since the remaining tags of images in Corel5k
are very sparse (less than 5), thus using tags as features would cause performance
deterioration.

\section{Experiments}
\label{sec:exp}
In this section, our experimental setup is first outlined, followed by the analysis of some
parameters. Finally, the performance of the proposed fomulation is evaluated and compared
with prior methods.
\subsection{Datasets and Measurement}
\label{ssec:exp.dataset}

To facilitate comparison between our method and previous ones, the same datasets
and features as in \cite{lin2013image} are used. Two datasets are used: the well-established benchmark
 dataset Corel5k and the real-world Flickr30Concepts. Statistics of both datasets
are given in Table \ref{table.dataset}.

For Corel5k dataset, 40\% of tags are randomly deleted which ensures
that each image has at least one tag removed and one tag remained. The 1000-dimensional
SIFT BoW feature is downloaded from
\url{http://lear.inrialpes.fr/people/guillaumin/data.php.} Random deletion is performed 8
times and averaged performance is reported. Furthermore, a validation set containing 491
images is extracted randomly to perform parameter tuning.

For Flickr30Concepts dataset \cite{lin2013image}, the data provided by the
authors are used, including the ground truth and the initial tagging matrix, along with two types
of features: the 1000-dimensional SIFT BoW feature and the composite features consisting of
a set of 10 kinds of basic features\footnote{The features include: Color Correlogram, Color Layout,
CEDD, Edge Histogram, FCTH, JCD, Jpeg Coefficient Histogram, RGB Color Histogram,
Scalable Color, SURF with Bag-of-Words model.}.

Also, the same test method as \cite{lin2013image} is used, as with the same measurements:
{\it average precision@N} ({\it i.e.} AP@N), {\it average recall@N} ({\it i.e.} AR@N)
and {\it coverage@N} ({\it i.e.} C@N). Evaluations are only conducted for the test set, and
extract neighbors only in the training set, for a fair comparison. \\
\begin{table}[t]
\centering
\caption{Statistics of Corel5k and Flickr30Concepts. Counts of tags are given in format of "mean/maximum".}
\label{table.dataset}
\begin{tabular}{l|l|l}
&Corel5k&Flickr30Concepts\\
\hline \hline
Vocabulary Size&260&2,513\\
\hline
Nr. of Images&4,918&27,838\\
\hline
Tags per Image&3.4/5&8.3/70\\
\hline
Del. Tags per Image&1.4 (40\%)&3.3 (40\%)\\
\hline
Test Set&492&2,807\\
\hline
\end{tabular}

\end{table}
\subsection{Parameter Settings}
\label{ssec:params}
Altogether 6 parameters are involved in the proposed method, hence it is necessary to
tune each parameter in order to achieve better performance and analyze their respective
influence to the completion process.

\begin{figure}[!t]
\begin{minipage}[t]{1\linewidth}
\includegraphics[width=1\textwidth]{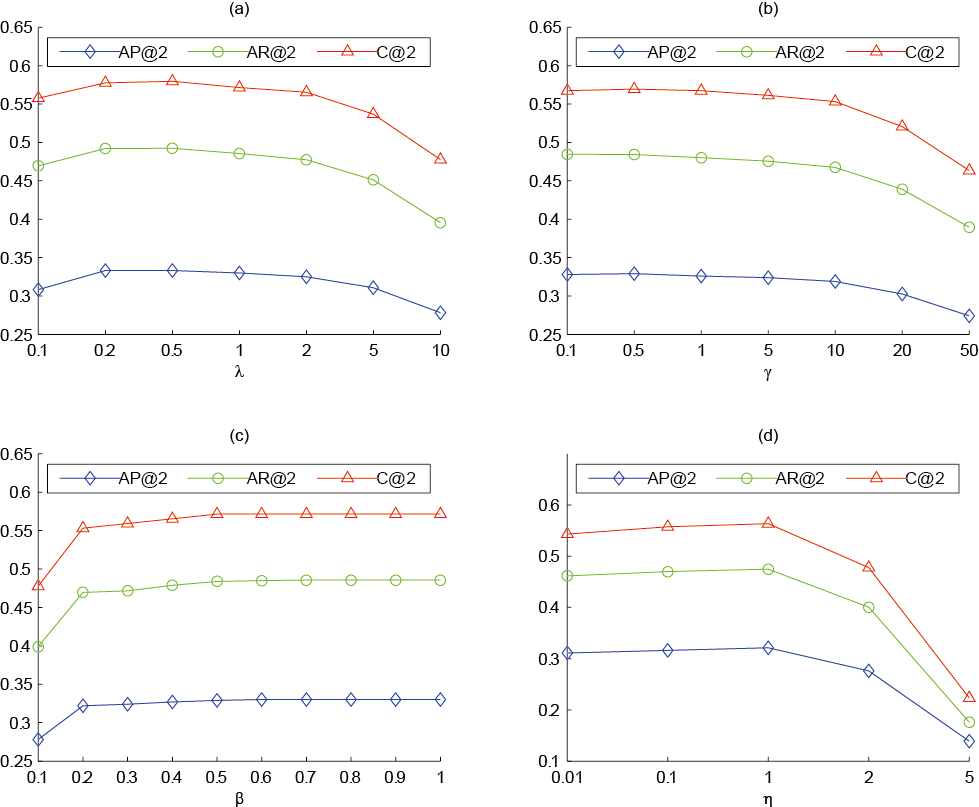}
\caption{Influences of $\lambda, \gamma, \beta$ and $\eta$ on validate set of Corel5k.}
\label{fig.params}
\end{minipage}
\end{figure}

The control variable method is adopted, which means modifying only one parameter at
a time and keeping others unchanged. The results are shown in Fig.\ref{fig.params}.
Since $\alpha$ and $\mu$ have little influence, so a large number (here 1) is used to make the feature-sign method faster.

As illustrated in Fig.1, the value of $\gamma$ should not be too large, since a larger
$\gamma$ means a higher degree of confidence on the assumption that $SA\sim A$, whereas
this maybe questionable due to the semantic gap. Similarly, performance also degraded as
$\lambda$ gets larger, since $T$ is obtained from incomplete initial tags. A smaller $\beta$
means a denser $E$, thus, as $\beta$ vanishes, it would be difficult to achieve fine-grained
reconstruction of $D$, so the completed tags may be inaccurate; On the other hand,
if $E$ gets too sparse, its ability to control the completion process would be weaken.
Here too large values are not used since the feature-sign method would return all-zero matrix
when $\beta$ gets too large. For $\eta$, as it approaches 0, the $L_{1}$ regularizer
seems disabled; as it gets larger, $V$ maybe too sparse and the reconstruction error would get
large.
The final values adopted in our experiments are $\lambda=0.5, \gamma=1, \beta=0.7, \eta=1$.
\subsection{Tag Completion Results}
\label{ssec:results}
To demonstrate the effectiveness of our method, its performance is compared with state-of-the-art
annotation methods (JEC \cite{makadia2008new} and TagProp \cite{guillaumin2009tagprop}) and
several newly proposed tag completion algorithms, namely TMC, DLC and LSR. Note that JEC and
TagProp are designed for multi-features, while TMC and DLC are more suitable for SIFT BoW
feature, whereas the LSR method, along with the proposed one, can handle both multi-features
and SIFT BoW feature. For these baseline methods,
the evaluation results reported in \cite{lin2013image} are directly cited. Experimental
results using only the SIFT BoW feature on both datasets are shown in Table \ref{table.sift},
and results using 10 kinds of features on Flickr30Concepts are presented in Table \ref{table.10fea}.

For Corel5k dataset, the proposed method outperforms previous methods by a large
margin, especially for DLC and TMC, which have been analyzed in Section \ref{sec:intro}.
Note that the pre-processing steps of obtaining $S$ and $T$ in our method correspond
to the LSR method, which is far more delicate in the design of group-sparsity
regularizer and soft fusion of coefficients. However, the LSR method is highly dependent
on initial labels, thus, if some critical tags are removed, the sparse reconstruction
may turn out inaccurate. Our method, on the other hand, seeks a balance between low-rank
completion and sparse reconstruction, thus its ability to recover latent data gets preserved.
\begin{table}[t]
\centering
\caption{Experimental results on Corel5k and Flickr30Concepts with only SIFT BoW feature.}
\label{table.sift}
\begin{tabular}{l|c|c|c|c|c|c}
&\multicolumn{3}{c}{Corel5k}&\multicolumn{3}{|c}{Flickr30Concepts}\\
&\multicolumn{3}{c}{($N=2$)}&\multicolumn{3}{|c}{($N=4$)}\\
&$AP$&$AR$&$C$&$AP$&$AR$&$C$\\
\hline \hline
TMC&0.23&0.33&0.40&0.19&0.21&0.37\\
\hline
DLC&0.09&0.13&0.18&0.07&0.09&0.23\\
\hline
LSR&$\mathbf{0.28}$&$\mathbf{0.42}$&$\mathbf{0.50}$&$\mathbf{0.30}$&$\mathbf{0.36}$&$\mathbf{0.60}$\\
\hline \hline
Ours&$\mathbf{0.32}$&$\mathbf{0.49}$&$\mathbf{0.57}$&$\mathbf{0.32}$&$\mathbf{0.39}$&$\mathbf{0.64}$\\
\hline
\end{tabular}

\end{table}

\begin{table}[t]
\centering
\caption{Experimental results on Flickr30Concepts with 10 types of features.}
\label{table.10fea}
\begin{tabular}{l|c|c|c}
&\multicolumn{3}{c}{Flickr30Concepts}\\
&\multicolumn{3}{c}{$(N=4)$}\\
&$AP$&$AR$&$C$\\
\hline \hline
JEC&0.25&0.30&0.49\\
\hline
TagProp&$0.23$&0.29&0.50\\
\hline
LSR&$\mathbf{0.37}$&$\mathbf{0.45}$&$\mathbf{0.67}$\\
\hline \hline
Ours&$\mathbf{0.39}$&$\mathbf{0.48}$&$\mathbf{0.72}$\\
\hline
\end{tabular}

\end{table}

For the Flickr30Concepts dataset, the increase in performance with respect to LSR is
not so significant as for Corel5k dataset, since images contained in Flickr30Concepts
have richer initial labels than images in the former dataset, thus the requirement for
robustness towards noisy initial tags is more essential for Corel5k. Note that JEC and
TagProp both perform tag propagation according to the similarities defined by distances
in feature space, thus all suffer from the problem which has been mentioned in Section
\ref{sec:intro}.

Finally, compared with results using only SIFT BoW feature, performances using 10 types
of features get substantially improved, both for the LSR method and the proposed one,
which once more demonstrates the superiority of multiple features.

\section{Conclusions}
\label{sec:conclusion}
A novel tag completion algorithm is proposed in this paper, which is characterized by
the low-rank, error sparsity, and the ability to preserve local linear reconstruction
structures in the compressed low-dimensional feature space and tag space. Extensive
experiments conducted on the well-known Corel5k dataset and the real-world
Flickr30Concepts dataset demonstrate the
effectiveness and efficiency of the proposed algorithm, where our method outperforms
prior methods by a large margin.
%

\bibliographystyle{IEEEbib}
\bibliography{refs}

\end{document}